\documentclass{article}

\usepackage[final]{corl_2020} 

\usepackage{multirow}
\usepackage{booktabs}
\usepackage{graphicx}
\usepackage{wrapfig}
\usepackage[font=small,labelfont=bf]{caption}
\usepackage{amsmath,amssymb} 
\usepackage[T1]{fontenc}
\usepackage{enumitem}

\newcommand{\tasknamenosp}{SPF}
\newcommand{\fulltasknamenosp}{Sequential Pointcloud Forecasting}
\newcommand{\methodnamenosp}{SPFNet}
\newcommand{\fullapproachnamenosp}{Sequential Pointcloud Forecasting for Sequential Pose Forecasting}
\newcommand{\approachnameacronymnosp}{SPF${}^2$}
\newcommand{\taskname}{\tasknamenosp~}
\newcommand{\methodname}{\methodnamenosp~}
\newcommand{\fulltaskname}{\fulltasknamenosp~}
\newcommand{\fullapproachname}{\fullapproachnamenosp~}
\newcommand{\approachnameacronym}{SPF${}^2$~}

\title{Inverting the Pose Forecasting Pipeline with SPF2: Sequential Pointcloud Forecasting for \\Sequential Pose Forecasting}

\author{
  Xinshuo Weng${}^{1}$, Jianren Wang${}^{1}$, Sergey Levine${}^{2}$, Kris Kitani${}^{1}$, Nicholas Rhinehart${}^{2}$ \\
  ${}^{1}$Robotics Institute, Carnegie Mellon University \\
  \texttt{\{xinshuow, jianrenw, kkitani\}@cs.cmu.edu} \\
  ${}^{2}$Berkeley Artificial Intelligence Research Lab, University of California, Berkeley \\
  \texttt{\{svlevine, nrhinehart\}@eecs.berkeley.edu} \\
}

\begin{document}
\maketitle


\vspace{-0.9cm}
\begin{abstract}
Many autonomous systems forecast aspects of the future in order to aid decision-making. For example, self-driving vehicles and robotic manipulation systems often forecast future object poses by first detecting and tracking objects. However, this detect-then-forecast pipeline is expensive to scale, as pose forecasting algorithms typically require labeled sequences of object poses, which are costly to obtain in 3D space. Can we scale performance without requiring additional labels? We hypothesize yes, and propose inverting the detect-then-forecast pipeline. Instead of detecting, tracking and then forecasting the objects, we propose to first forecast 3D sensor data (e.g., point clouds with $100$k points) and then detect/track objects on the predicted point cloud sequences to obtain future poses, i.e., a forecast-then-detect pipeline. This inversion makes it less expensive to scale pose forecasting, as the sensor data forecasting task requires no labels. Part of this work's focus is on the challenging first step -- Sequential Pointcloud Forecasting (SPF), for which we also propose an effective approach, SPFNet. To compare our forecast-then-detect pipeline relative to the detect-then-forecast pipeline, we propose an evaluation procedure and two metrics. Through experiments on a robotic manipulation dataset and two driving datasets, we show that SPFNet is effective for the SPF task, our forecast-then-detect pipeline outperforms the detect-then-forecast approaches to which we compared, and that pose forecasting performance improves with the addition of unlabeled data. Our project website is \textcolor{blue}{\url{http://www.xinshuoweng.com/projects/SPF2}}.
\end{abstract}

\vspace{-0.3cm}

\vspace{-0.45cm}
\section{Introduction \label{sec:intro}}
\vspace{-0.3cm}

Many autonomous systems forecast aspects of the future to aid decision-making. For example, self-driving vehicles \cite{Liang2020,Zeng2019,Wang2018} and robotic systems \cite{Byravan2017,Manglik2019} often forecast future object poses using pose forecasting algorithms \cite{Lan2014,Koppula2016,byravan2017se3, Byravan2018,Rhinehart2018,Gupta2018,Kosaraju2019,Rhinehart2019,Chang2019}. Given sensor data, these approaches first detect/track objects, and then train a pose forecasting algorithm to predict future poses, e.g., object location and velocity. However, this detect-then-forecast pipeline is expensive to scale, as pose forecasting algorithms typically require labels of future object poses and their correspondences in time, which are costly to obtain, especially in 3D space \cite{Caesar2019}. Can we create a pose forecasting pipeline that is cheaper to scale by leveraging large unlabeled 3D point cloud sequences?

We hypothesize yes, and propose an inverted forecast-then-detect pipeline (point cloud forecasting followed by detection/tracking). By reserving the order, forecasting is now performed on the point clouds, which does not require any expensive-to-collect labels. Notably, the first step in this pipeline is a challenging problem that no prior work has addressed (the closest is perhaps \cite{byravan2017se3}, which predicts flow of the points and requires point-wise annotations). Therefore, we consider this first step an important research direction in its own right, and believe it may prove useful to other downstream tasks. We term this task \fulltaskname (\tasknamenosp): given a sequence of past point clouds as input, output a sequence of future point clouds. Two benefits of the \taskname task are that (1) the prediction represents the entire scene, which contains information about both scene background and foreground objects, and (2) it requires only point cloud data, which costs significantly less to collect than labels, and is already prevalent in public datasets \cite{Chang2019,Caesar2019,Geiger2012}. 

Towards the \taskname task, we present \methodnamenosp, which employs an LSTM autoencoder model. To leverage 2D Convolutional Neural Networks (CNNs) while preserving 3D structure of the point cloud, we use a range map representation \cite{Caccia2019}. For training and evaluation, we use raw LiDAR point clouds from two real-world autonomous driving datasets (KITTI \cite{Geiger2012} and nuScenes \cite{Caesar2019}) and depth point clouds from a Baxter robot dataset \cite{byravan2017se3}. We show that \methodname can reliably forecast future point clouds in different domains and outperform competitive approaches.

We then present our forecast-then-detect pipeline, \approachnameacronymnosp, which reverses the steps in the conventional detect-then-forecast pipeline that first performs perception (3D object detection \& tracking) and then trajectory forecasting. Specifically, our \approachnameacronym pipeline first performs forecasting at the sensor data level using \methodnamenosp, and then uses off-the-shelf 3D detection/tracking on the predicted point clouds to obtain future object trajectories. This pipeline inherits the benefits of the \taskname task: (1) forecasting at the sensor data level can leverage the geometry of the scene -- when forecasting point clouds of the entire scene, locations of every point are implicitly constrained by those of surrounding points; (2) our forecasting module is unsupervised, which facilitates training with data at larger scales than supervised pose forecasting. Note that the detector/tracker in our pipeline may still require labels.

Standard trajectory forecasting evaluation assumes access to the set of ground-truth (GT) past trajectories, and each GT past trajectory is used as input to forecast a future trajectory. As the correspondence between each GT past and GT future trajectory is known, every predicted trajectory has a target GT future trajectory. This \emph{a priori} knowledge of the mapping between each forecasted trajectory and its target GT future trajectory is required to compute the conventional error metrics (ADE and FDE \cite{alahi2016}). However, come deployment time, both the conventional detect-then-forecast pipeline and the \approachnameacronym pipeline do not have access to GT past trajectories. As a result, there is no known correspondence between each forecasted and GT trajectory, which prevents compututation of the ADE/FDE metrics. To evaluate detect-then-forecast and our pipeline in an end-to-end fashion, we propose a procedure that first establishes an association between forecasted and GT trajectories with a matching step. Unlike standard ADE/FDE computation, this association does not provide $100$\% recall of the GT trajectories -- some GT trajectories may be missed, with true positives identified by thresholding the ADE score. Because different methods may achieve different recalls, we propose metrics to \emph{average} ADE/FDE over recalls. Our contributions are summarized as follows: 
\vspace{-2mm}
\begin{enumerate}[leftmargin=1cm]
    \itemsep0em 
    \item \textbf{A novel forecast-then-detect pipeline, \approachnameacronymnosp}, that can scale forecasting performance by leveraging unlabeled point cloud data and outperform the detect-then-forecast pipeline.
    \item \textbf{A new task, \fulltaskname (\tasknamenosp)}, which predicts a 3D representation of the future of the entire scene and does not require any human annotation for training.
    \item \textbf{An effective approach for \tasknamenosp}, deemed \methodnamenosp, that outperforms competitive approaches in one robot manipulation dataset and two real-world autonomous driving datasets.
    \item \textbf{An evaluation procedure with two metrics} that can evaluate performance of the detect-then-forecast pipeline and our \approachnameacronym pipeline in an end-to-end fashion.
\end{enumerate}


\vspace{-0.45cm}
\section{Related Work}
\vspace{-0.3cm}

\noindent\textbf{Forecasting Tasks.} Prior tasks can be categorized into trajectory forecasting, activity forecasting and video forecasting. Trajectory (or pose) forecasting \cite{Gupta2018,Deo2018,Kosaraju2019,Rhinehart2019,Ivanovic2019,Weng2020_GNNTrkForecast,Liang_2020_CVPR} predicts a sequence of future ground positions of target objects. Activity forecasting \cite{Lan2014,Koppula2016} predicts a person's future actions (such as walking and running). Both trajectory forecasting and activity forecasting are object-level forecasting, which (1) require object labels; (2) solely predict information about target objects while ignoring everything else in the scene. In contrast, our \taskname task predicts a richly informative 3D representation of the entire scene and does not require any human annotation for object labels.

Video forecasting \cite{Jing2018,Oliu2018} is similar to \tasknamenosp, as both tasks forecast high-dimensional sensor data. The difference is that video forecasting outputs a geometrically-projected 3D representation, whereas \taskname outputs a full 3D representation. Beyond forecasting RGB images, \cite{Qi2019,Mahjourian2017} jointly predict the depth image and RGB image in the next frame, which is also related to \tasknamenosp, as depth forecasting is a 3D forecasting task, although the data representation is different and \cite{Qi2019,Mahjourian2017} only perform one-frame prediction. More importantly, prior 3D forecasting methods have primarily focused on the accuracy of the forecasting task itself, while we focus both on the accuracy of the \taskname task and on utilizing it to enhance performance of the downstream pose forecasting task.

\noindent\textbf{Point Cloud Generation.} Prior work focuses on generating a single object point cloud. \cite{Fan2017} proposes to use Chamfer distance \cite{Fan2017} and Earth Mover’s distance \cite{Rubner2000} as loss functions to generate object point clouds. \cite{Lin2018} generates object point cloud and then project into multiple 2D viewpoints optimized with a 2D re-projection error. Beyond deterministic point cloud generation, prior work employs stochastic frameworks such as flow-based models \cite{yang2019}, generative adversarial networks (GANs) \cite{Shu2019} and variational autoencoders (VAEs) \cite{Klokov2019}. 

Although most prior work focuses on point cloud generation for a single object, a few works attempted to generate point clouds for an entire scene, which we call scene point clouds. \cite{Caccia2019} uses VAEs and GANs to generate a scene point cloud from a random noise vector. \cite{Tomasello2019} generates a high-resolution scene point cloud conditioned on stereo images and low-resolution LiDAR inputs. \cite{byravan2017se3} predicts a scene point cloud in the next frame by considering the action applied to an object in the scene. Different from \cite{Caccia2019,Tomasello2019,byravan2017se3} which output a single scene point cloud, \taskname generates a sequence of scene point clouds. To the best of our knowledge, \taskname is the first formulated as a sequence prediction of large-scale scene point clouds ($100$k points per frame).

\noindent\textbf{Perception and Trajectory Forecasting Pipeline}\label{sec:related}
aims to localize objects and predict where they will move in the future, given the sensor data. The conventional detect-then-forecast pipeline has two modules: (1) a perception module (including detection \cite{Lee2016,Luo2018,Weng20192,Wang2020_GNNDetTrk,Weng2018_R2N} and tracking \cite{Weng2019,Weng2020_GNN3DMOT,Weng2020_eccvw}) to obtain past object trajectories; (2) a trajectory forecasting module to predict the future object trajectories based on the past object trajectories. Prior work often separately evaluates perception and forecasting, e.g., using average precision and CLEAR metrics \cite{Bernardin2008} for perception evaluation, and use ADE and FDE \cite{alahi2016} to evaluate trajectory forecasting. However, modularized trajectory forecasting evaluation lacks the ability to assess the performance of the entire pipeline, as it employs the unrealistic assumption that we have access to past GT object trajectories.

Different from the conventional perception and forecasting pipeline, our pipeline first performs the forecasting at the sensor level and then applies detection and tracking to the predicted sensor data to obtain forecasted pose trajectories. Also, our new evaluation procedure can account for the error from detection, tracking and forecasting modules, enabling the evaluation over the entire pipeline. Although \cite{Liang_2020_CVPR} also propose to evaluate the entire pipeline in an end-to-end fashion, their evaluation is performed at a single operating point, while our evaluation with two proposed metrics can summarize performance over many operating points.


\begin{figure}[t]
    \centering
    \includegraphics[trim=0cm 8.7cm 1.7cm 0cm, clip=true, width=\linewidth]{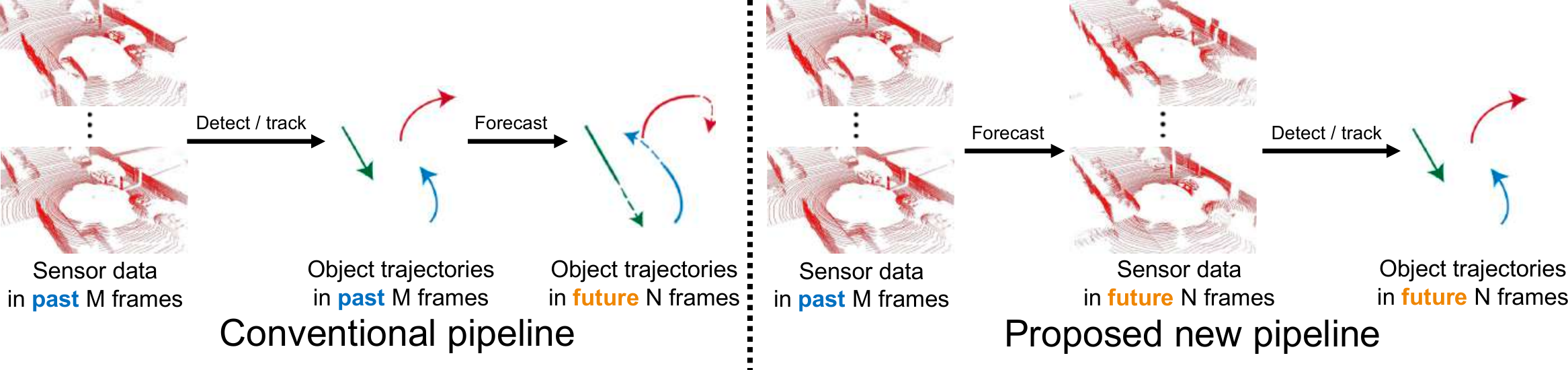}
    \vspace{-0.75cm}
    \caption{\small{\textbf{Comparison of perception and trajectory forecasting pipeline.} (Left) Conventional detect-then-forecast pipeline. (Right) Our pipeline first forecasts point clouds, and then detects and tracks objects.}}
    \label{fig:pipeline}
    \vspace{-0.4cm}
\end{figure}

\vspace{-0.4cm}
\section{The \approachnameacronym Pipeline\label{sec:pipeline}}
\vspace{-0.3cm}

As mentioned in Sec.~\ref{sec:related}, the conventional pipeline is separated into two steps as in Fig. \ref{fig:pipeline} (left): (1) a detection/tracking module to obtain past object trajectories; (2) a forecasting module to predict future trajectories based on the past trajectories. In contrast, we propose a new pipeline, \fullapproachname (\approachnameacronymnosp), depicted in Fig. \ref{fig:pipeline} (right). Specifically, we first forecast point cloud sequences using our \methodname (Sec. \ref{sec:spcsfnet}), and then use a 3D detector and tracker on the predicted point cloud sequences to obtain pose sequences (trajectories). 


\vspace{-0.35cm}
\subsection{\fulltasknamenosp}
\vspace{-0.2cm}

\textbf{Task Formulation.} We first define a 3D point cloud enclosing an entire scene as the scene point cloud, which includes points belonging to both objects and scene background. The goal of \taskname is to predict a sequence of future scene point clouds given a sequence of past scene point clouds. Specifically, given $M$ frames of past scene point clouds with each frame $S_t = \{(x, y, z)_j\}_{j=1}^{K_t}$, where $t \in [-M+1, \cdots\!, 0]$ denotes the frame index, $j \in [1, \cdots\!, K_t]$ denotes the index of points and $K_t$ denotes the number of points at frame $t$, the goal of \taskname is to predict $N$ frames of future scene point clouds with each frame $S_t = \{(x, y, z)_j\}_{j=1}^{K_t}$, where $t \in [1, \cdots\!, N]$. 


\vspace{-0.35cm}
\subsection{Approach: \methodname \label{sec:spcsfnet}}
\vspace{-0.2cm}

\begin{wrapfigure}{r}{0.65\linewidth}
\vspace{-0.1cm}
    \centering
    \includegraphics[width=\linewidth]{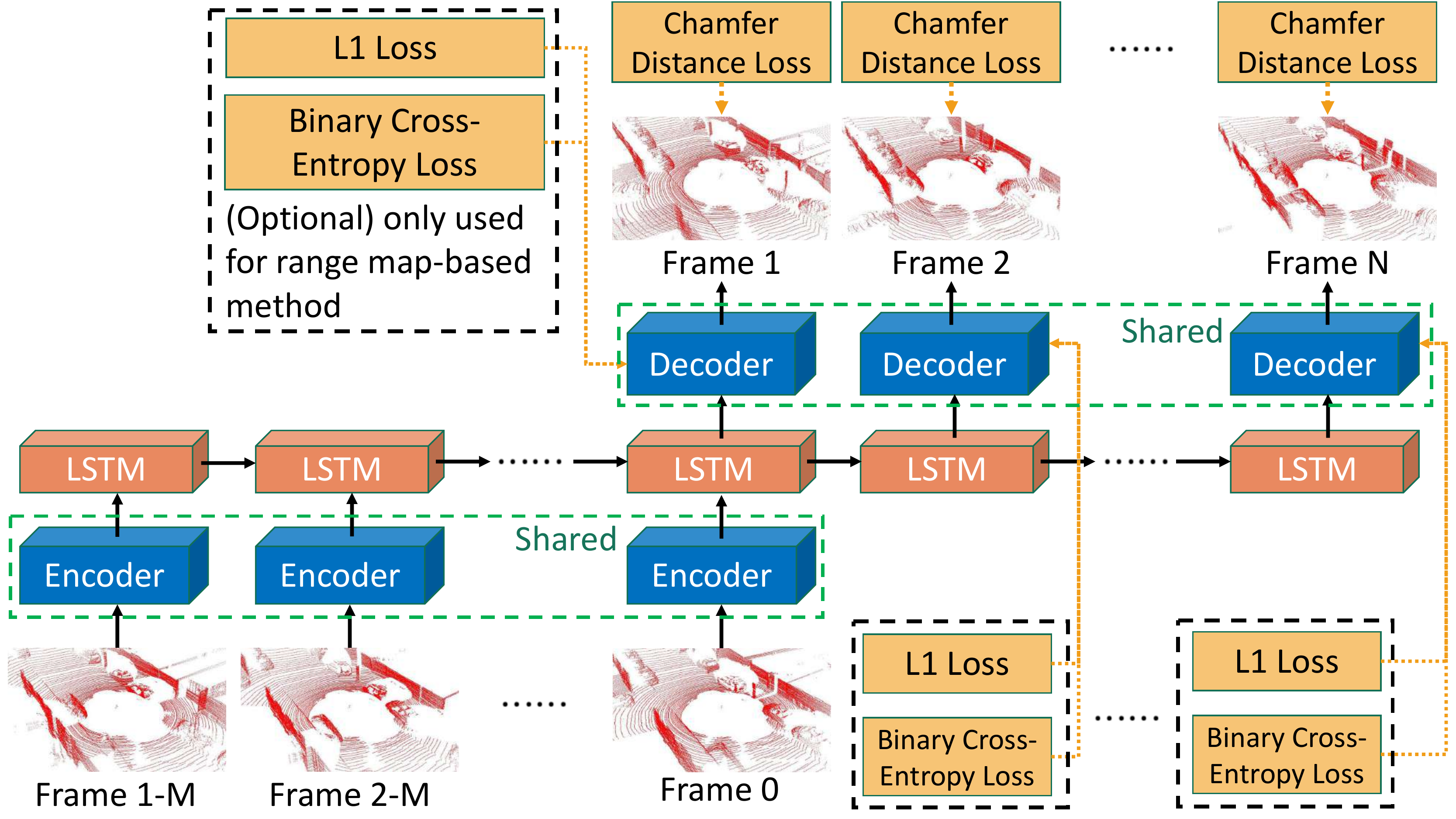}
    \vspace{-0.6cm}
    \caption{Proposed \methodnamenosp.
    }
    \label{fig:main}
    \vspace{-0.45cm}
\end{wrapfigure}

Our \methodname is shown in Fig. \ref{fig:main}, which has four modules: (a) a shared encoder to extract features from past scene point clouds, (b) an LSTM to model temporal dynamics of the features of the scene point clouds, (c) a shared decoder to generate scene point clouds in the future and (d) a combination of the L1 and binary cross-entropy losses is applied to the generated intermediate range maps and a Chamfer distance loss is applied to the predicted point clouds at every future frame. 

\noindent\textbf{Encoder.} To predict the future scene point clouds, it is important to extract useful information from the past through an encoder. In this work, we investigate two types of encoders:

\textit{(1) Point-based encoder.} To process a point cloud, PointNet \cite{qi2016} is a straightforward choice. We use a network (multi-layer perceptron (MLP) plus a max-pooling operator) similar to \cite{qi2016} to obtain the local feature for each point and the global feature for the scene. To leverage both the local per-point feature and global scene feature, we concatenate them so that every point has the feature of its own and the global feature. However, preserving high-dimensional features for each point will reach the GPU memory limit (e.g., RTX 2080 Ti has 11Gb), as \taskname requires to process a sequence (e.g., 30 frames) of point clouds each with a large number of points (e.g., $100$k). In fact, if maintaining high-dimensional features for each point, the required GPU memory will be massive. As a result, we add another MLP and a max-pooling operator to obtain a single compact feature to represent the entire scene point cloud at every frame as the output of the encoder.

\textit{(2) Range map-based encoder.\label{sec:rangemap}} In addition to directly extract the feature from point clouds, another popular approach in the literature \cite{Liang2018,Caccia2019} is to first transform the point cloud into a 2D representation and then use 2D CNNs for feature extraction. Although these methods are fast and efficient, the disadvantage of most of these methods (except for the range map \cite{Caccia2019}) is to lose significant information. For example, if projecting the point cloud to the bird's eye view, information in the height direction is lost. As a result, we choose to use the range map as our 2D representation \cite{Caccia2019}, which preserves full 3D information in the LiDAR point cloud. For details about how to obtain the range map from the input scene point cloud, please refer to our supplementary materials. After the scene point cloud is transformed into a 2D range map with a size of $H \times W$ (we use $120 \times 1024$ in our experiments to predict $122.88$k points per frame), we use a standard 2D CNN with $8$ layers (convolution + batch norm + ReLU) to extract features. For the detailed network architecture of our encoders, we provide visualization in our supplementary. 

\noindent\textbf{Temporal Dynamics Modeling.} As \taskname is formulated as a sequence-to-sequence problem, we use a standard LSTM network to learn temporal dynamics of the features. Specifically, we feed the feature obtained from the encoder in the past $M$ frames to the LSTM network and predict features for the future $N$ frames in an autoregressive manner.

\noindent\textbf{Decoder.} To predict future scene point clouds, a decoder is needed to convert the predicted features from LSTM to scene point clouds. With two encoders carefully designed, the corresponding decoder follows the encoder design in a symmetric manner: (1) For the point-based decoder, we first increase the dimensionality of the LSTM output feature to $1 \times (K \times 3)$ using MLP, and then reshape it to a scene point cloud with a size of $K \times 3$. (2) For the range map-based decoder, we use a 2D CNNs with eight layers (deconv + batch norm + ReLU) to map the LSTM predicted feature back to a range map with $H \times W$. Additionally, to deal with the holes in the range map (\emph{i.e.,} the pixels that do not have any point in the point cloud), we also predict a binary range mask $R_M$ ($0$ denotes pixel without a point) with a shape of $H \times W$ using a similar 2D CNNs. Then, when we transform the predicted range map back to a scene point cloud only for the pixels that are not masked out.

\noindent\textbf{Losses.} We use the Chamfer distance \cite{Fan2017} loss $\mathcal{L}_{cd}$ to minimize the distance between ground truth $S^*_{1:N}$ and predicted point cloud sequences $\hat S_{1:N}$. For the range map-based method, we apply two additional losses on the intermediate range map outputs to form our full objective in Eq.~\ref{eq:objective}: (1) a $L_1$ distance loss between the ground truth, $R^*$, and predicted, $\hat R$, 2D range maps; (2) a binary cross-entropy loss $L_{bce}$ between the predicted $\hat R_M$ and ground truth $R_M^*$ range mask. 
\vspace{-0.15cm}
\begin{align}
    &\mathcal{L} \doteq \sideset{}{_{i=1}^{N}} \sum \mathcal{L}_{cd}(\hat S_i, S_i^*) + \lambda_1 \mathcal{L}_{1}(\hat R_i, R_i^*) + \lambda_2 \mathcal{L}_{bce}(\hat {R_M}_i, {R_M}^*_i),\label{eq:objective}\\
   &\mathrm{where}~\mathcal{L}_{cd}(\hat S, S^*)\doteq\sideset{}{_{x\in\hat S}} \sum\min_{y\in S^*}\|x-y\|^2_2 + \sideset{}{_{y\in S^*}} \sum \min_{x\in S^*}\|x-y\|^2_2, \nonumber
\end{align}

\vspace{-0.5cm}
\subsection{3D Object Detection and 3D Multi-Object Tracking}
\vspace{-0.2cm}

Once we predict future point clouds using \methodnamenosp, the last step of our \approachnameacronym pipeline is 3D detection and tracking to obtain objects' future poses. In theory, our \approachnameacronym pipeline is generic so any detector and tracker can be used, e.g., using unsupervised detectors/trackers to make our full pipeline unsupervised. However, existing unsupervised 3D detectors often have lower accuracy than supervised 3D detectors. To ensure accuracy of our \approachnameacronym pipeline, we use a supervised 3D detector \cite{Shi2019} while use an unsupervised 3D tracker \cite{Weng2019}, which can achieve strong tracking performance in KITTI. 

\vspace{-0.4cm}
\section{End-to-End Perception and Trajectory Forecasting Evaluation}
\vspace{-0.3cm}

\begin{figure}[t]
    \centering
    \includegraphics[trim=0cm 8.5cm 9cm 0cm, clip=true, width=0.62\linewidth]{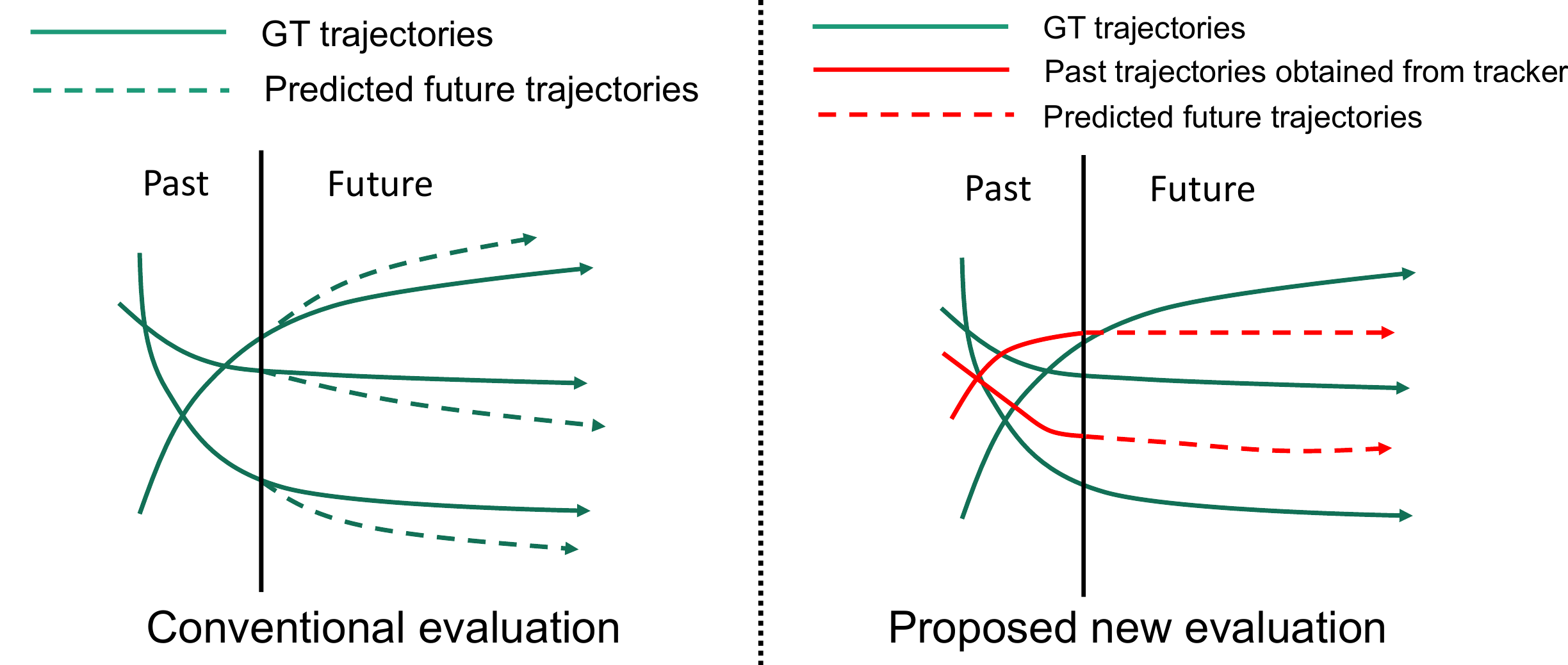}
    \includegraphics[trim=0.5cm 0.3cm 0.4cm 0.5cm, clip=true, width=0.35\linewidth]{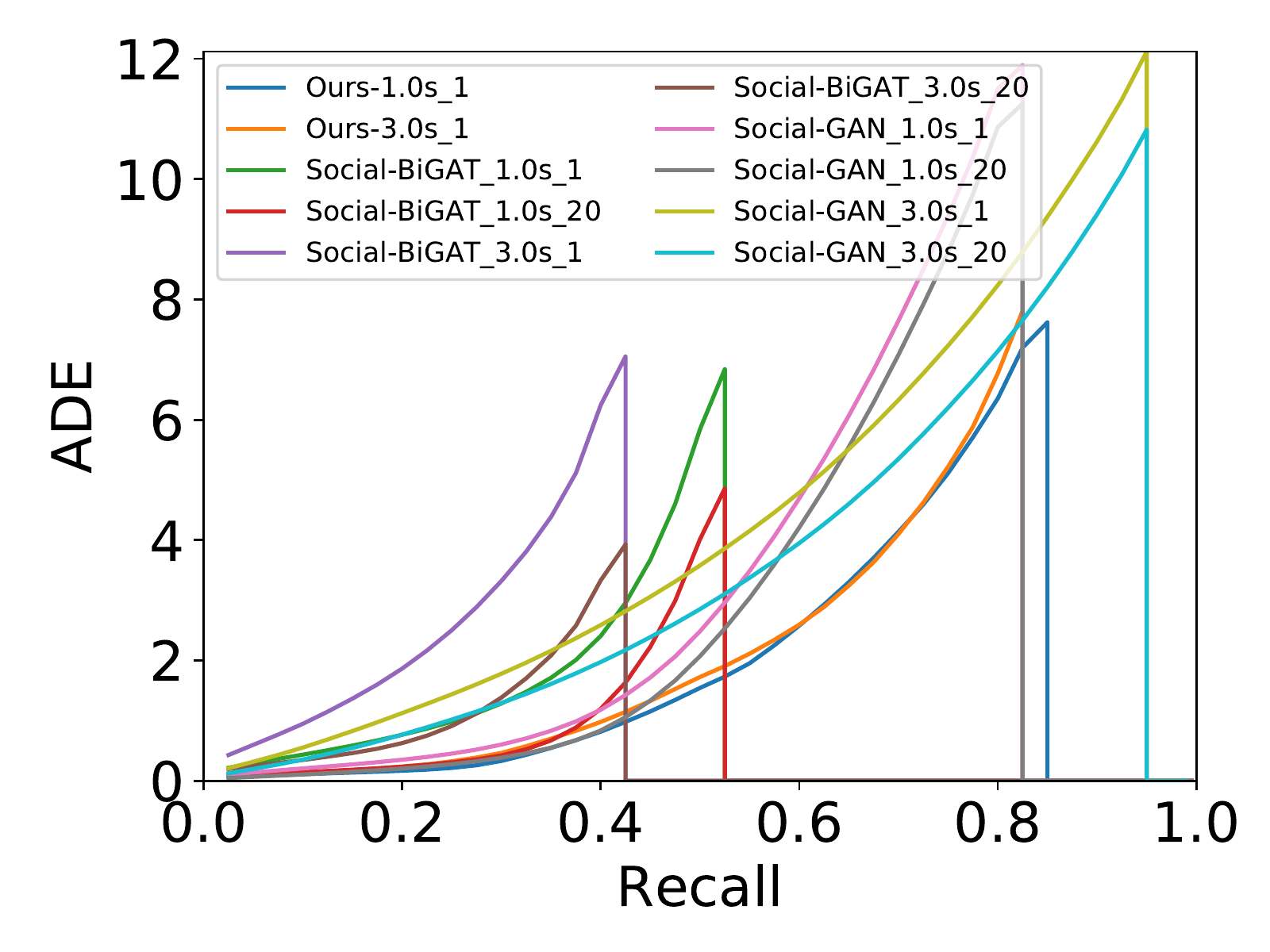}
    \vspace{-0.3cm}
    \caption{\textbf{(Left) Comparison of the evaluation procedure.} Conventional evaluation uses GT past to predict the future, which has the prior knowledge of the mapping between each predicted and GT future trajectory. In the real world, people use tracking outputs as past trajectories so the predicted trajectories do not have known mapping with GT, and thus a matching step is needed. \textbf{(Right) ADE-over-recall curve.} Methods can have different maximum recalls so it is not fair to compare ADE of each method performing at its maximum recall. Therefore, integral metrics that summarize performance across the same range of recalls are preferred. 
    }
    \label{fig:evaluation}
    \vspace{-0.7cm}
\end{figure}

As mentioned in Sec. \ref{sec:intro} and \ref{sec:related}, standard modularized evaluation for trajectory forecasting relies on the unrealistic assumption that we have access to GT past trajectories. As a result, standard modularized evaluation is not applicable for evaluating either the detect-then-forecast or our pipeline. Consider two common forecasting metrics: Average Displacement Error (ADE) and Final Displacement Error (FDE) \cite{alahi2016}, which compute the distance between each pair of prediction and its corresponding GT. In the modularized evaluation setting, as GT past trajectories are used, computing ADE and FDE is straightforward as each predicted future trajectory has its corresponding future GT trajectory. However, in the end-to-end evaluation setting, computing ADE and FDE requires an additional matching step between the forecasted trajectories and GT trajectories as their correspondences are not known as \emph{a priori}: see Fig. \ref{fig:evaluation} (left). To obtain the correspondences, we match forecasted trajectories to GT trajectories based on ADE scores using the Hungarian algorithm \cite{WKuhn1955}. Please refer to our supplementary material for details regarding this matching step.

As errors can be accumulated in detection/tracking/forecasting, the final forecasted trajectories are not perfect and some GT objects might not be matched to any predicted trajectory (i.e., a false negative). We identify true positives by thresholding the ADE score. In general, the recall value will not be $100\%$ and it might vary substantially across methods. This is in contrast to the modularized evaluation where the recall is always $100\%$. As an example, we show the ADE-over-recall curve computed on the KITTI dataset in Fig. \ref{fig:evaluation} (right) for our method and two baselines (see Sec. \ref{sec:trajectory_eval} for details). We can see that ADE tends to increase as recall increases, i.e., a looser ADE threshold will include more high-ADE forecasted trajectories. Therefore, to achieve a relatively fair comparison, methods must compare ADE either at an identical recall or by averaging ADE across the same range of recalls. We choose the latter solution because evaluating at a single recall can produce less robust numbers\footnote{\cite{Liang_2020_CVPR} identify true positives using a 3D IoU threshold of 0.5, i.e., a single operating point. As a result, the reported numbers do not account for performance at other operating points, e.g., at a higher recall including objects below the IoU threshold of 0.5, e.g., those difficult to detect.}, e.g., some methods might perform better at a recall point but worse at other recalls. To that end, we propose two metrics called AADE/AFDE (\underline{A}verage ADE/FDE) to summarize ADE/FDE performance across a range of recalls. Specifically, AADE/AFDE are computed by integrating ADE/FDE. Similar to other integral metrics such as average precision in object detection, we approximate the integration by a summation of the ADE/FDE over a set of recalls. 

\vspace{-0.4cm}
\section{Experiments}
\vspace{-0.3cm}

\noindent\textbf{Datasets.} 
We evaluate on two real-world autonomous driving datasets: KITTI-Raw \cite{Geiger2012}, nuScenes \cite{Caesar2019}, and a robot manipulation dataset of the Baxter robot \cite{Byravan2018}. 

\vspace{-0.3cm}
\subsection{Evaluating \methodname in the Autonomous Driving Domain}
\vspace{-0.2cm}


\textbf{Evaluation and Settings.} 
We use standard Earth Mover's Distance (EMD) \cite{Rubner2000} and Chamfer Distance (CD) \cite{Fan2017} to measure the distance between GT and predicted future point clouds. We evaluate two settings for all methods: (1) observe $1.0$s past and predict the next $1.0$s; (2) observe $3.0$s past and predict the next $3.0$s. For KITTI, $1.0$s and $3.0$s means $10$ and $30$ frames. For nuScenes key-frames only, $1.0$s and $3.0$s means $2$ and $6$ key-frames.
 
\noindent\textbf{Baselines.}
As this paper is the first towards the \taskname task, there are no existing methods for comparison. Therefore, we devise baselines based on existing techniques: (1) \textit{Identity}. We duplicate the scene point cloud from the last past frame to the future for $N$ times; (2) \textit{GT-Ego}. We use ground truth ego-motion (rotation and translation) between adjacent past frames and compute an average motion. Then, we use the average motion to warp the last past frame of point cloud to the future frames; (3) \textit{Est-Ego}. Instead of using the ground truth ego-motion, we estimate the ego-motion using \cite{Zhou2017}; (4) \textit{Align}. We use the point cloud alignment methods (ICP \cite{Besl1992} and Deep-ICP \cite{Wang2019}) to compute the global rigid motion between adjacent past frames and obtain the average motion for warping; (5) \textit{SceneFlow}. We use scene flow methods \cite{Liu2019} to estimate the point-wise motion between the last two past frames and use the motion to warp every single point. 

\begin{table}[t]
\begin{center}
\caption{Quantitative evaluation for the proposed \taskname task on the KITTI and nuScenes datasets.}
\resizebox{\textwidth}{!}{
\begin{tabular}{@{}llrrrrrrrr@{}}
\toprule
Datasets \ \ \  \ & Metrics \ \ & \ \ Identity & \ \ GT-Ego & \ \ Est-Ego & \ \ Align-ICP & \ \ Align-\cite{Wang2019} & \ \ SceneFlow & \ \ \textbf{Ours+Point} & \ \ \textbf{Ours+RM} \\ 
\midrule

\multirow{2}{*}{KITTI-1.0s} & CD$\downarrow$ & 12.82 & 5.47 & 9.18 & 6.13 & 6.02 & 3.15 & 2.37 & 0.92 \\
& EMD$\downarrow$ & 526.87 & 391.03 & 495.21 & 418.25 & 439.17 & 291.63 & 211.47 & 128.81 \\ 
\multirow{2}{*}{KITTI-3.0s} & CD$\downarrow$ & 13.31 & 7.91 & 11.31 & 9.14 & 9.57 & 5.08 & 3.91 & 1.57 \\
& EMD$\downarrow$ & 602.89 & 452.81 & 502.83 & 470.25 & 493.26 & 351.46 & 267.42 & 175.54 \\

\midrule
\multirow{2}{*}{nuScenes-1.0s} & CD$\downarrow$ & 8.42 & 2.16 & 4.91 & 4.04 & 3.50 & 1.93 & 1.25 & 0.40 \\
& EMD$\downarrow$ & 461.63 & 168.37 & 299.13 & 281.53 & 270.81 & 117.41 & 135.94 & 78.37 \\ 
\multirow{2}{*}{nuScenes-3.0s} & CD$\downarrow$ & 10.16 & 2.85 & 6.52 & 7.13 & 5.27 & 3.81 & 2.97 & 0.69 \\
& EMD$\downarrow$ & 494.81 & 190.14 & 370.91 & 419.37 & 332.97 & 294.53 & 128.26 & 91.83 \\

\bottomrule
\label{tab:spcvf} 
\end{tabular}}
\end{center}
\vspace{-1.2cm}
\end{table}

\noindent\textbf{Results.} We summarize the results in Table \ref{tab:spcvf}, where +RM and +Point denote our range map-based and point-based method. Our \methodname consistently outperforms baselines in both metrics. The results are reasonable because our \methodname (1) is designed for sequence prediction and (2) can learn the future location for each individual point, while the baselines are weak because they are originally designed for one-frame prediction (i.e., warping between two frames) and also most baselines only estimate global motion. However, it is a common difficulty to find strong baselines when proposing a new task. Also, we observed that Ours+RM performs better than Ours+Point. We hypothesize that predicting in the range map space may be relatively easier than in the point cloud space for our \taskname task, although it is possible to see further improvements for the point-based method if we use a more advanced point-based encoder. Additionally, we find that the results on nuScenes seem to consistently better than KITTI, which we believe are primarily due to two reasons: 1) nuScenes LiDAR sensor has a smaller sensing range (e.g., 70m v.s. 120m of the KITTI LiDAR sensor), and forecasting for the nearby range is generally considered easier; 2) nuScenes keyframes have a lower frame rate (2Hz v.s. 10Hz in KITTI), leading to fewer frames of prediction at the same time horizon.

\begin{wrapfigure}{r}{0.5\linewidth}
    \centering
    \vspace{-0.55cm}
    \includegraphics[trim=1.1cm 0.4cm 0.2cm 0.4cm, clip=true, width=0.49\linewidth]{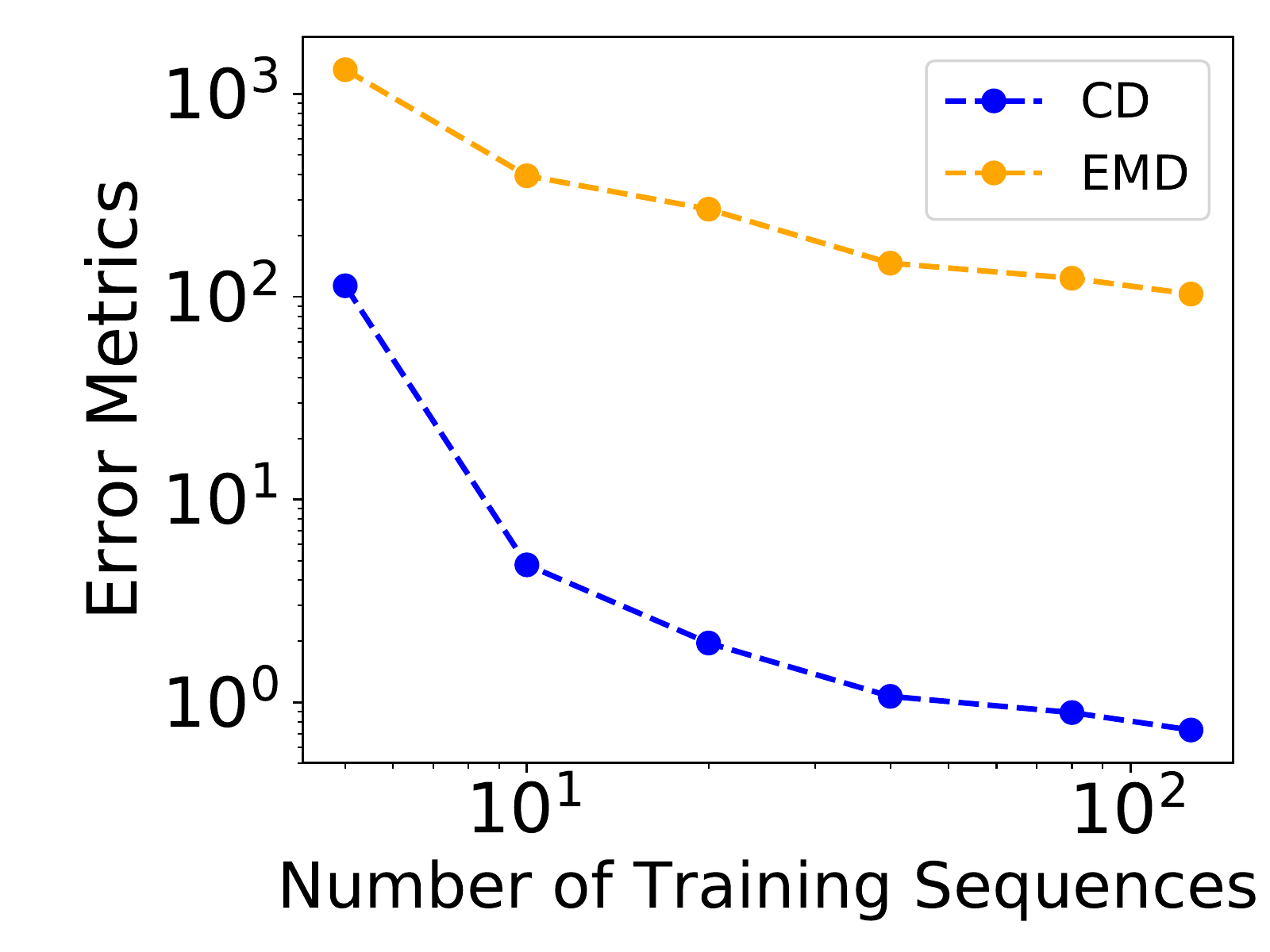}
    \includegraphics[trim=1.1cm 0.4cm 0.6cm 0.4cm, clip=true, width=0.49\linewidth]{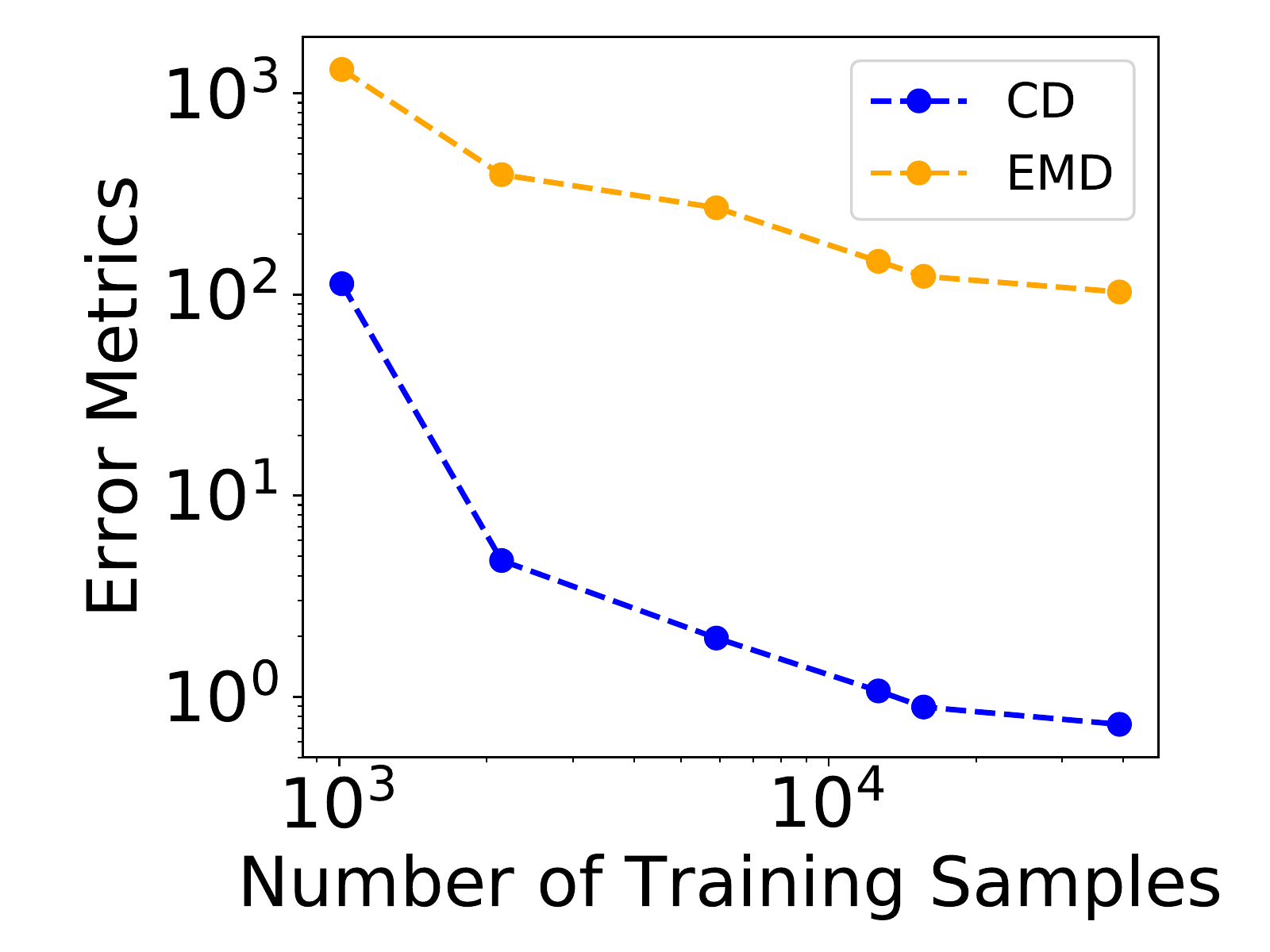}
    \vspace{-0.6cm}
    \caption{Error v.s. amount of training data.}
    \label{fig:scaling}
    \vspace{-0.6cm}
\end{wrapfigure}

\noindent\textbf{Performance scaling with unlabeled data.} As our \methodname can be trained with massive unlabeled point clouds without any human annotation, it is useful to know if its performance scales with the amount of training data. To that end, we train our \methodname on the KITTI dataset multiple times, each time a different number of training sequences are withheld. This experiment is conducted in the 1.0s prediction setting using the range map-based method. The results are shown in Fig. \ref{fig:scaling} with two sub-figures, one with number of training samples and the other with number of training sequences as the $x$ axis. Clearly, with more training sequences or more training samples, CD and EMD are decreasing, which confirms our hypothesis that it is cheaper to scale performance of the proposed \methodname by leveraging unlabeled point cloud data.


\begin{table}[t]
\begin{center}
\caption{Quantitative evaluation on the Baxter dataset. Note that the numbers within the parentheses for the PPFE metric are what we have reproduced, which are very close to the reported results by the authors of the baselines. This demonstrates that we successfully reproduced the baselines for comparison.}
\resizebox{\textwidth}{!}{
\begin{tabular}{@{}lrrrrr@{}}
\toprule
Metrics \ & \ \ SE3-Pose-Nets + Joint Angles  & \ \ SE3-Pose-Nets \cite{Byravan2018} & \ \ SE3-Nets \cite{Byravan2017} & \ \ \textbf{Ours+Point} & \ \ \textbf{Ours+RM} \\ 
\midrule

PPFE$\downarrow$    & 0.017 (0.019) & 0.027 (0.025) & 0.022 (0.024) & 0.035 & 0.030 \\ 
CD$\downarrow$      & 0.315         & 0.313         & 0.357         & 0.153 & 0.094 \\ 
EMD$\downarrow$     & 52.7          & 50.9          & 64.1          & 33.5  & 18.2  \\ 

\bottomrule
\label{tab:baxter} 
\end{tabular}}
\end{center}
\vspace{-1.1cm}
\end{table}

\vspace{-0.3cm}
\subsection{Evaluating \methodname in the Robot Manipulation Domain}
\vspace{-0.2cm}

\begin{wrapfigure}{r}{0.3\linewidth}
\centering
\vspace{-1cm}
\includegraphics[trim=0cm 0cm 0cm 0cm, clip=true, width=\linewidth]{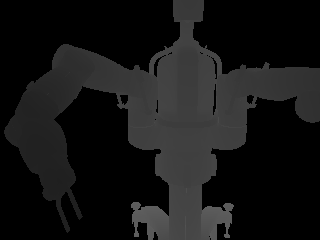}
\vspace{-0.5cm}
\caption{A example depth image of the Baxter robot.}
\label{fig:baxter}
\vspace{-0.4cm}
\end{wrapfigure}

To show that our proposed \methodname can be applicable to not only the autonomous driving domain but also other domains such as robot manipulation, we evaluate on the Baxter dataset collected by \cite{Byravan2018} using the Baxter simulator. Specifically, the dataset contains a Baxter robot (shown in Figure \ref{fig:baxter}) which moves its right arm with seven joints towards the target position in front of an RGBD camera.

\noindent\textbf{Evaluation Metrics.} 
As the Baxter dataset provides camera matrix and pixel correspondences between depth images across frames, computing flow in 3D space is possible. Therefore, we follow \cite{Byravan2018} and report a per-point flow error (PPFE) to measure the differences between predicted and ground truth flow in the 3D space, in addition to the EMD and CD metrics.
 
\noindent\textbf{Baselines.} We compare against S.O.T.A. methods (SE3-Nets \cite{Byravan2017}, SE3-Pose-Nets \cite{Byravan2018}, and its one variant). Note that these baselines are different from our \methodname in two-fold: 1) they only support one-frame prediction (i.e., not sequence-to-sequence forecasting) and it is nontrivial to adapt them for sequence forecasting. Thus, to obtain a direct comparison, we only test the first predicted frame of our \methodnamenosp, though we train our \methodname with 10 input and 10 output frames to better learn scene dynamics; 2) instead of predicting the next frame of the point cloud, these baselines predict the flow of each point and then warp the past point cloud to the next frame with the predicted flow. As a result, these baselines require point correspondences to obtain ground truth flow for training and cannot be applied to LiDAR point clouds which generally do not have point correspondences. 

\noindent\textbf{Results.}
We summarize the results in Table \ref{tab:baxter}. Note that only results for the PPFE metric is reported in \cite{Byravan2018}. In order to compute the results of baselines for CD and EMD, we first need to re-train the baselines (no pre-trained model is released by the authors of the baselines). To make sure that we successfully re-trained the baselines, we show the reproduced results on the PPFE metric within the parentheses, which are very close to the reported results in \cite{Byravan2018}. In fact, our re-trained SE3-Pose-Net obtained a slightly lower error in FFPE than the reported number in \cite{Byravan2018}, i.e., 0.025 v.s. 0.027. With successfully re-trained baselines, we can now additionally compute for CD and EMD metrics. We can see that,  our \methodname achieved much better performance on the CD and EMD metrics than all baselines. This is probably because the baselines are trained to optimize the flow instead of the point cloud distance (e.g., CD or EMD). On the other hand, through our \methodname is not optimized for the flow, our FFPE is competitive to the baselines such as SE3-Pose-Net (0.030 v.s., 0.027). Overall, the results show that our \methodname is able to learn scene dynamics on a robot dataset.


\vspace{-0.3cm}
\subsection{Evaluating Perception and Trajectory Forecasting Pipeline\label{sec:trajectory_eval}}
\vspace{-0.2cm}

\noindent\textbf{Evaluation and Settings.} 
We use the new evaluation procedure with the AADE/AFDE metrics. To fairly compare AADE/AFDE for all methods, we only integrate ADE/FDE up to a recall that all methods can reach. This is because we cannot integrate ADE/FDE for a method beyond its maximum recall, which is undefined. We use the $1.0$ and $3.0$ seconds prediction. As trajectory forecasting modules used in the conventional pipeline often perform stochastic prediction, we evaluate them by sampling $1$ and $20$ times. For our pipeline with deterministic forecasting module \methodnamenosp, it can be viewed as $1$ sample. We use the range map-based encoder here for our method. 

\noindent\textbf{Baselines.} We employ different trajectory forecasting modules \cite{Gupta2018,Kosaraju2019,Chandra2019,Deo2018} in the conventional perception and trajectory forecasting pipeline. For a fair comparison, we use the same detector \cite{Shi2019} and tracker \cite{Weng2019} for both our and conventional pipeline. Although \cite{Gupta2018,Kosaraju2019} are originally proposed for predicting the trajectory of people, we adapt them to work with the car. 

\begin{table}[t]
\begin{center}
\caption{Evaluation for the perception and trajectory forecasting pipeline on the KITTI and nuScenes datasets.}
\vspace{0.05cm}
\resizebox{\textwidth}{!}{
\begin{tabular}{@{}llrrrrrr@{}}
\toprule
Datasets \ & Metrics \ & \ Samples & \ Conv-Social \cite{Deo2018} & \ Social-GAN \cite{Gupta2018} & \ Social-BiGAT \cite{Kosaraju2019} & \ TraPHic \cite{Chandra2019} & \ \ \  \textbf{Ours} \\ 
\midrule

\multirow{4}{*}{KITTI-1.0s} & \multirow{2}{*}{AADE$\downarrow$} 
                                    &  1 & 0.792 & 0.524 & 1.099 & 0.470 & 0.317 \\
                                  & & 20 & 0.623 & 0.340 & 0.443 & 0.382 & --  \\
& \multirow{2}{*}{AFDE$\downarrow$} &  1 & 1.285 & 0.886 & 1.708 & 0.889 & 0.405 \\
                                  & & 20 & 1.152 & 0.511 & 0.546 & 0.613 &  -- \\ 
\midrule

\multirow{4}{*}{KITTI-3.0s} & \multirow{2}{*}{AADE$\downarrow$} 
                                    &  1 & 1.692 & 1.362 & 2.720 & 1.432 & 0.408  \\ 
                                  & & 20 & 1.593 & 0.984 & 1.231 & 0.725 & --  \\
& \multirow{2}{*}{AFDE$\downarrow$} &  1 & 2.670 & 2.267 & 3.938 & 2.536 & 0.504 \\
                                  & & 20 & 2.385 & 1.512 & 1.405 & 1.118 & --  \\
\midrule

\multirow{4}{*}{nuScenes-1.0s} & \multirow{2}{*}{AADE$\downarrow$} 
                                    &  1 & 1.186 & 1.117 & 2.030 & 1.214 & 0.821 \\ 
                                  & & 20 & 0.907 & 0.762 & 0.826 & 0.881 & --  \\
& \multirow{2}{*}{AFDE$\downarrow$} &  1 & 1.490 & 1.310 & 2.337 & 1.563 & 0.825 \\
                                  & & 20 & 1.231 & 0.763 & 0.849 & 1.197 & --  \\
\midrule

\multirow{4}{*}{nuScenes-3.0s} & \multirow{2}{*}{AADE$\downarrow$} 
                                    &  1 & 1.794 & 2.224 & 4.954 & 2.417 & 1.044 \\
                                  & & 20 & 1.658 & 1.426 & 1.760 & 1.938 &  --  \\
& \multirow{2}{*}{AFDE$\downarrow$} &  1 & 2.850 & 3.224 & 6.765 & 3.479 & 1.043 \\
                                  & & 20 & 2.538 & 1.652 & 1.845 & 2.766 & --  \\
\bottomrule
\label{tab:traj} 
\end{tabular}}
\end{center}
\vspace{-1.1cm}
\end{table}

\noindent\textbf{Results.} We summarize performance on the KITTI and nuScenes datasets in Table \ref{tab:traj}. We can see that our \approachnameacronym outperforms the conventional detect-then-forecast pipeline using different forecasting modules with $1$ sample. When sampling $20$ times for stochastic forecasting modules in the conventional pipeline, we do see improvements, but surprisingly, our deterministic \approachnameacronym with $1$ sample still achieves the best performance in most settings. We hypothesize it is because that (1) our forecasting at the sensor level is indeed effective in our \approachnameacronym pipeline, and (2) the data used for evaluation might not contain much uncertainty, i.e., future is quite certain as vehicles are moving straight most of the time. As a result, evaluation on other datasets with diverse futures might be valuable in the future.


\begin{figure}[t]
    \centering
    \vspace{0.23cm}
    \includegraphics[trim=0cm 7cm 0cm 0cm, clip=true, width=\textwidth]{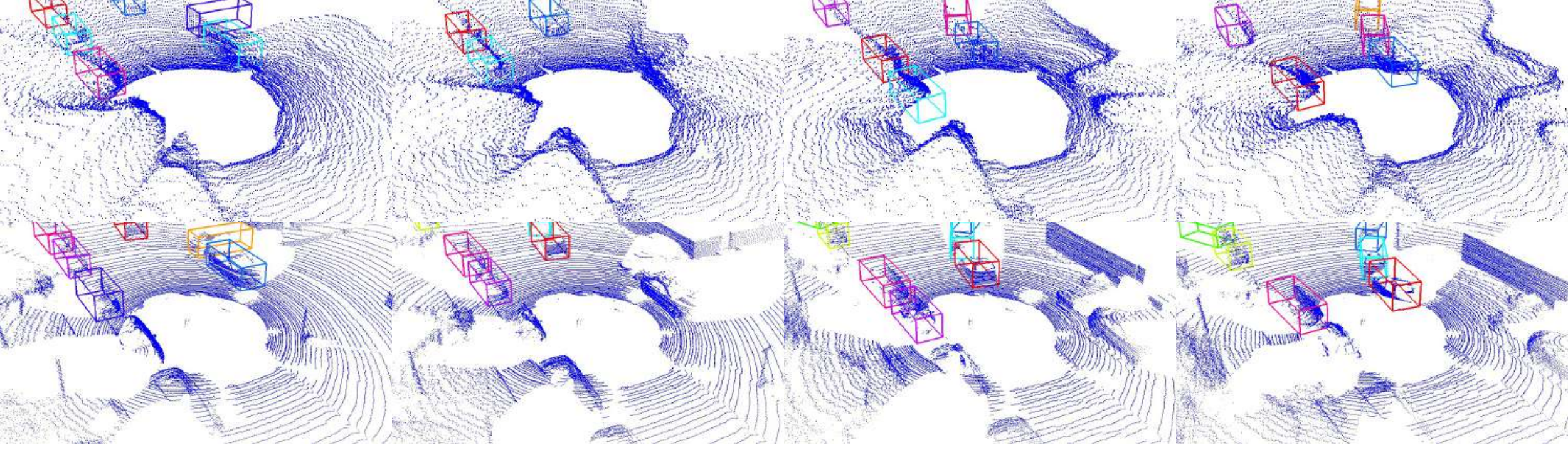}
    \vspace{-0.6cm}
    \caption{Qualitative comparison between our prediction and GT on KITTI. \textbf{(Top)} Predicted future scene point clouds by our \methodname and detection/tracking outputs on the predicted point clouds by our \approachnameacronym pipeline. Bounding boxes with the same color across frames represent a predicted object trajectory by our pipeline; \textbf{(Bottom)} GT future point clouds and GT bounding boxes. Note that KITTI only provides object labels in the frontal view, \emph{i.e.}, objects that are behind the ego-vehicle do not have bounding box labeled. Also, the color of the same object might be different in prediction and GT as their absolute IDs (identities) are different.}
    \label{fig:spcvf_qua}
    \vspace{-0.6cm}
\end{figure}

\noindent\textbf{Visualization.} We visualize the results of our \methodname on the KITTI dataset for the 3.0s prediction setting in Fig. \ref{fig:spcvf_qua} (top). We plot $4$ frames of results from a prediction of $30$ frames with an interval of $7$ frames, showing that our predicted scene point clouds reasonably reconstruct the objects and scene, and align well with GT shown in Fig. \ref{fig:spcvf_qua} (bottom). However, the current limitation is that the shape of the objects in predicted point clouds are not sharp enough. Moreover, we visualize detection/tracking outputs (colored bounding boxes) on top of our predicted scene point clouds, showing that the tracking outputs of our \approachnameacronym pipeline are very close to the GT bounding boxes.

\vspace{-0.3cm}
\section{Conclusion}
\vspace{-0.25cm}

We proposed a new perception and trajectory forecasting pipeline, \approachnameacronymnosp, that inverts the detect-then-forecast pipeline. Towards our \approachnameacronym pipeline, we proposed a new forecasting task, \tasknamenosp, and method, \methodnamenosp, which predicts a richly informative 3D representation of the future and does not need any human annotation for training. Also, an evaluation procedure is proposed to support end-to-end evaluation of the perception and forecasting pipeline. We verified the performance of our \methodname and \approachnameacronym pipeline on two real-world autonomous driving datasets and one robotic manipulation dataset, which demonstrates that our forecasting performance can scale by leveraging unlabeled data and the proposed \taskname is useful to the downstream task of trajectory forecasting. As our inverted pose forecasting pipeline is generic, it might be possible to extend it to other instances of pose forecasting (not only trajectory forecasting) domains, such as future activity forecasting. 



\clearpage
\acknowledgments{This research was supported in part by the Office of Naval Research and ARL DCIST CRA W911NF-17-2-0181, and JST AIP Acceleration, Grant Number JPMJCR20U1, Japan.
}


\vspace{-0.1cm}
\bibliography{main}  

\end{document}


\maketitle

\vspace{-0.8cm}
\section{Overview}
\vspace{-0.2cm}

As it is difficult to include every detail of our work in the main paper, we provide further details here about our new pipeline \approachnameacronym, the proposed \taskname task, the first approach \methodnamenosp, proposed evaluation procedure for the pipeline, and additional experiments to support our main paper.

\vspace{-0.3cm}
\section{New Perception and Trajectory Forecasting Pipeline: \approachnameacronymnosp}
\vspace{-0.2cm}

\vspace{-0.2cm}
\subsection{Fine-tuning of the Object Detector}
\vspace{-0.1cm}

As generated scene point clouds from \methodname are not exactly same as the ground truth scene point clouds, using 3D detector pre-trained on the ground truth scene point clouds in our \approachnameacronym pipeline can lead to inferior performance. To adapt the 3D detector to work with our generated scene point clouds and achieve higher performance, we fine-tune the 3D detector using the scene point clouds generated by our \methodname along with the 3D bounding box annotations. 

\vspace{-0.3cm}
\section{Our Proposed Task: \fulltaskname (\tasknamenosp)}
\vspace{-0.2cm}
\subsection{Scene Point Cloud v.s. Object Point Cloud}
\vspace{-0.1cm}

\begin{wrapfigure}{r}{0.47\linewidth}
\centering
\vspace{-0.9cm}
\includegraphics[trim=2cm 8.5cm 11.5cm 0cm, clip=true, width=\linewidth]{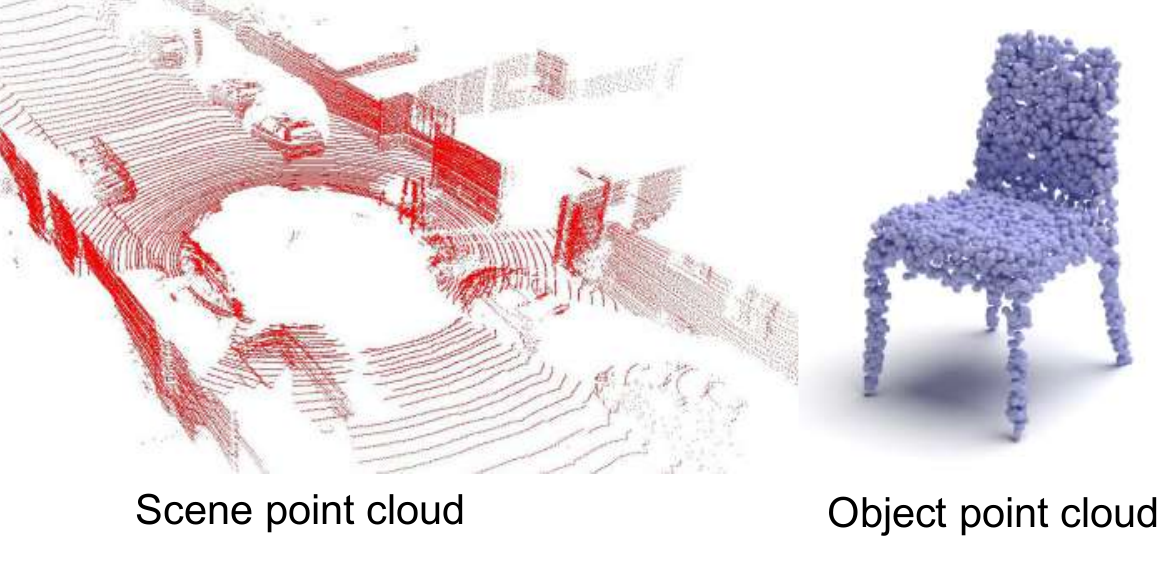}
\vspace{-0.8cm}
\caption{\small{Comparison between a scene point cloud ($100k$ points) and an object point cloud ($1k$ points).}}
\label{fig:spc}
\vspace{-0.3cm}
\end{wrapfigure}

We have defined the scene point cloud in Sec 3.1 of the main paper. Here, we visually compare the scene point cloud with the significantly different object point cloud in prior work. As an example, we show a scene point cloud captured by a LiDAR sensor from the KITTI dataset in Fig. \ref{fig:spc} (left) and an object point cloud in ShapeNet \cite{Chang2015} in Fig. \ref{fig:spc} (right). Obviously, scene point cloud contains much more information than the object point cloud, \emph{e.g.}, many instances of foreground objects such as cars, pedestrians, and background objects such as trees, fences, building, traffic sign that will affect the motion of foreground objects. As a result, it is much more challenging to model and generate scene point clouds than generate the object point cloud. Additionally, it is usually more difficult to process a scene point cloud with about $100k$ points than an object point cloud with about $1k$ points.

\vspace{-0.2cm}
\subsection{Relation to Object Trajectory Forecasting}
\vspace{-0.1cm}

\begin{figure}
\begin{center}
\includegraphics[trim=0.2cm 2.5cm 2.5cm 0cm, clip=true, width=\linewidth]{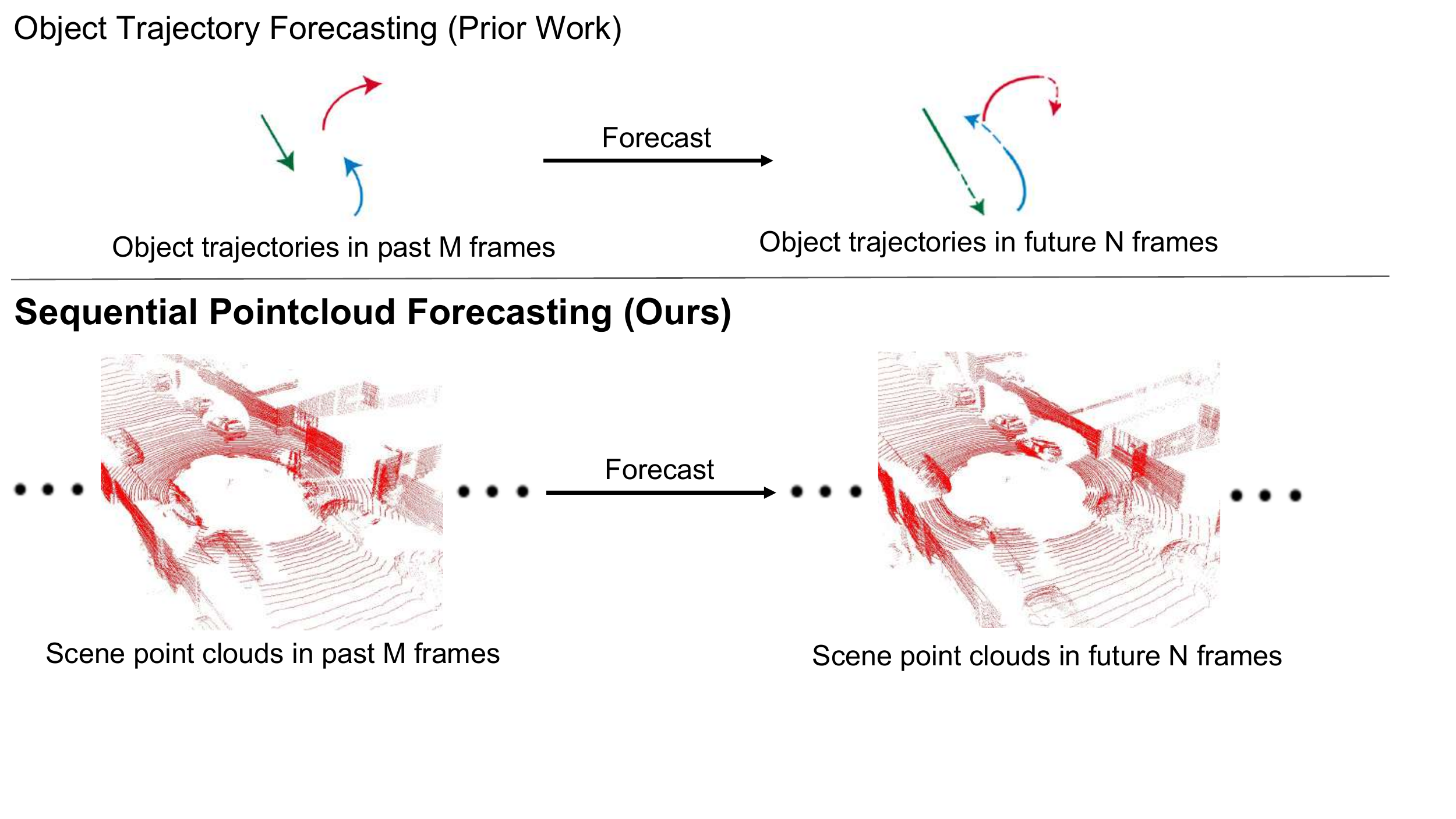}\\
\vspace{-0.2cm}
\caption{\textbf{(Top) Object Trajectory Forecasting}: given object trajectories in the past $M$ frames, the goal is to predict object trajectories in the future $N$ frames. Each colored arrow denotes an object's trajectory (from a top-down view). Solid parts of the arrows denote past trajectories and dotted parts denote future trajectories. \textbf{(Bottom) \fulltasknamenosp}: given 3D scene point clouds in the past $M$ frames, the goal is to predict a sequence of future scene point clouds.}
\label{fig:teaser}
\vspace{-0.3cm}
\end{center}
\end{figure}

In Sec. 2 of the main paper, we have described the differences between prior object trajectory forecasting task and our \tasknamenosp. Here, we additionally visualize the differences in Fig. \ref{fig:teaser}. Intuitively, the distinction lies in that, object trajectory forecasting aims to forecast future locations of a few points, where each point represents an object of interest, while our proposed \taskname task aims to predict future locations of all points in the scene point clouds, including points belong to the foreground objects and scene background. 

\vspace{-0.2cm}
\subsection{Relation to Scene Flow}
\vspace{-0.1cm}

Although scene flow \cite{Liu2019, Mayer2016} is not a forecasting task, scene flow task aims to estimate the 3D motion of each point in the scene point cloud, which is also related to our proposed \tasknamenosp. Here, we clarify the differences between scene flow and our \taskname to avoid any confusion. Specifically, the differences are three-fold: (1) training scene flow methods requires ground truth annotation of point-wise motion, which is expensive to obtain. In contrast, our proposed \taskname task only requires the scene point clouds (\emph{e.g.}, captured by a LiDAR sensor) as the ground truth for training and does not need to involve human annotation; (2) scene flow aims to learn point-wise motion between only two frames while our \taskname aims to learn scene dynamics from a sequence of point clouds (\emph{e.g.}, $30$ frames). As a result, \taskname is more challenging but can provide more useful information by modeling a history of past frames; (3) scene flow aims to learn motion in the current frame while \taskname is a forecasting task and aims to predict points' locations in the future (\emph{e.g.}, upto three seconds).

\vspace{-0.3cm}
\section{Approach: \methodnamenosp}
\vspace{-0.2cm}

\subsection{Preliminaries about the 2D Range Map Representation\label{sec:rangemap}}
\vspace{-0.1cm}

We briefly summarize the process of transforming a scene point cloud to a 2D range map here but refer the readers to \cite{Caccia2019} for details. Intuitively, range map transformation leverages the ray-casting nature of the LiDAR point cloud and projects the 3D points onto a 2D spherical plane. Specifically, for a 3D point $P=(x,y,z)$ in the scene point cloud with $z$ represents the height, its coordinate in the 2D spherical plane $Q = (\theta, \phi)$ is defined as the following: 
\begin{equation}
    \theta=\operatorname{atan} 2(y, x), \ \ \ \ \ \ \ \ \  
    \phi=\arcsin (z / \sqrt{x^{2}+y^{2}+z^{2}}),
\end{equation}
where $\theta$ is the azimuth angle and $\phi$ is the elevation angle from the origin. Then, to convert the continuous 2D spherical plane into a discrete 2D range map with resolution of $H\times W$, we discretize the 2D spherical plane with a fixed number of bins ($H$ bins along elevation axis, $W$ bins along azimuth axis) in a pre-defined range, \emph{e.g.}, $\phi \in [-30^{\circ}, 10^{\circ}]$ and $\theta \in [0^{\circ}, 360^{\circ}]$. The range is chosen as most points captured by a standard LiDAR sensor (\emph{e.g.}, Velodyne-64) fall into it. As a result, each bin (\emph{i.e.}, an pixel in the range map) will cover the range of $(\frac{40^{\circ}}{H}, \frac{360^{\circ}}{W})$ in the spherical plane, and we can easily find the nearest bin for each point $Q = (\theta, \phi)$ as its pixel location. In order to recover the 3D coordinate of a point from its 2D spherical coordinate, knowing the azimuth and elevation angle is not enough as they only provide the direction of the point to the origin in 3D space. Therefore, we use the distance $d = \sqrt{x^{2}+y^{2}+z^{2}}$ as the pixel value for each projected point in the 2D range map. In the case where multiple 3D points are projected to the same bin in the range map\footnote{Note that this situation happens very rarely in our experiments because we use a high-resolution range map so that there is no downsampling effect during the transformation of a point cloud to a 2D range map.}, we pick the one with the largest distance $d$ as the pixel value. This decision is made to preserve information in the faraway region of the point cloud, because there are often dense and redundant points in the nearby range but sparse points in the faraway range. For the bins that do not have any projected point, we simply fill their pixel value with zero and will mask them out during training and inference.  As a result, the range map with a resolution of $H \times W$ still preserves the 3D information and can be easily converted back to a point cloud. Note that the range map is different from a depth map in that (1) two maps are not in a same 2D coordinate; (2) depth map needs the camera matrix in order to be converted to a point cloud while the range map can be directly converted to a point cloud; (3) depth map usually only covers the points in frontal view while the range map can be defined to have the horizontal field of view of $360^{\circ}$ degrees.

\begin{figure}[t]
    \centering
    \includegraphics[trim=0cm 9cm 2.2cm 0cm, clip=true, width=\linewidth]{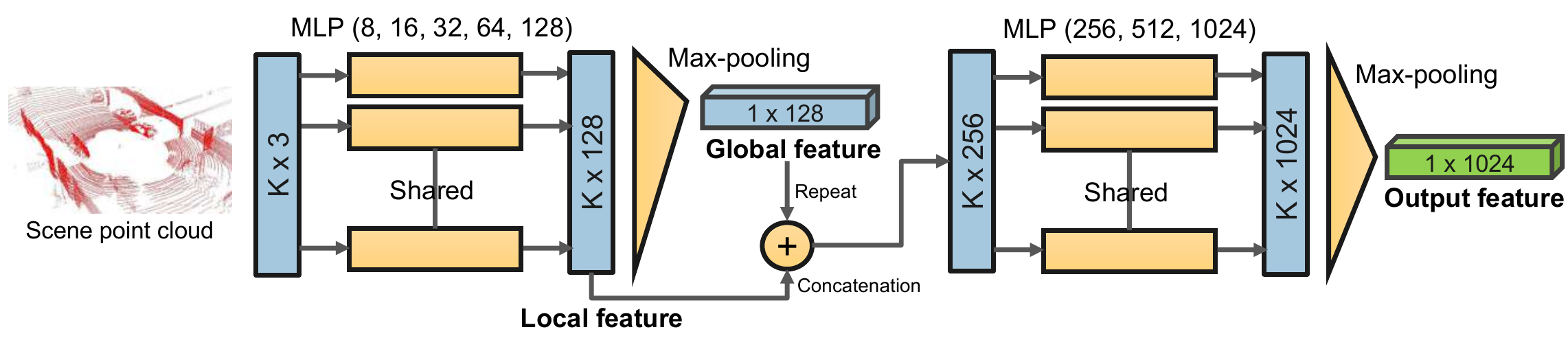} \\
    \includegraphics[trim=0cm 9.3cm 4.2cm 0cm, clip=true, width=\linewidth]{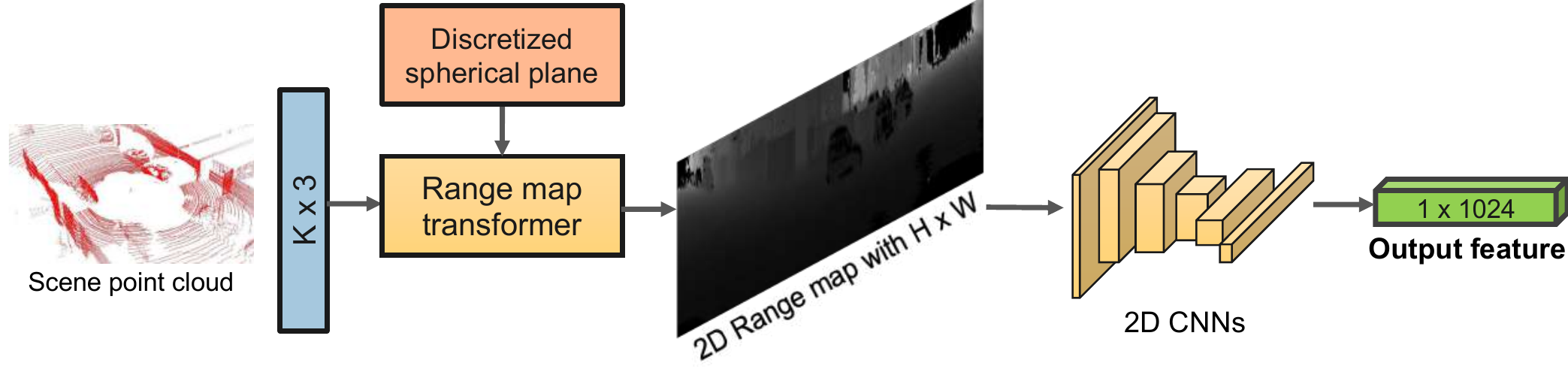}
    \vspace{-0.6cm}
    \caption{\textbf{(Top)}: Point-based encoder. \textbf{(Bottom)}: Range map-based encoder. Note that we only show one frame of the scene point cloud to be processed by the encoder in the above figure but our full \methodname uses encoder with shared weights to process a sequence of scene point clouds.}
    \label{fig:encoder}
    \vspace{-0.3cm}
\end{figure}

\vspace{-0.2cm}
\subsection{Detailed Network Architecture of the Encoder}
\vspace{-0.1cm}

We have briefly described our two (point-based and range map based) encoders in Sec. 3.2 of the main paper. To help readers better understand our \methodnamenosp, here we additionally illustrate the network architectures of our two encoders in Figure \ref{fig:encoder} and also provide more details of the neural network architectures below:

\textbf{Point-based encoder.} A scene point cloud with a shape of $K\times 3$ is fed into a shared five-layer multi-layer perceptron (MLP) (3$\Rightarrow$8$\Rightarrow$16$\Rightarrow$32$\Rightarrow$64$\Rightarrow$128) to obtain a local feature with a dimension of $128$ for each point. Then a global feature with the dimension of $1 \times 128$ is obtained by applying a max-pooling operator to the local features of all points. We then fuse the local and global features by concatenation to obtain a feature with a dimension of $256$ for each point. The fused features are then processed by a subsequent three-layer MLP (256$\Rightarrow$256$\Rightarrow$512$\Rightarrow$1024) and a max-pooling operator to obtain the output feature with a dimension of $1024$ for each point. As $K$ is very large (\emph{e.g.}, $>100k$) for every frame, maintaining high-dimensional features for every single point in the scene point cloud will reach the memory limit of the GPUs. As a result, we compress the features by a max-pooling operator to obtain the final output feature with a dimension of $1024$ to represent the entire scene point cloud.

\textbf{Range map-based encoder.} An input scene point cloud with a shape of $K \times 3$ along with a discretized spherical plane is fed into the range map transformer (see Sec. \ref{sec:rangemap}) to obtain the 2D range map with resolution of $H \times W$ (we use 120 x 1024 in our experiments to predict about 122.88k points per frame, similar to the number of points in the scene point cloud captured by a standard Velodyne LiDAR sensor). Then, we use a standard 2D CNNs with eight layers (each layer has convolution + batch normalization + ReLU) to extract the output feature from the 2D range map. Specifically, the eight layers in the 2D CNNs convert the input range map from a shape of $1\times120\times1024$, to $4\times60\times512$, to $8\times60\times256$, to $16\times60\times128$, to $32\times30\times64$, to $64\times15\times32$, to $128\times8\times16$, to $256\times4\times8$, to $512\times2\times4$. Then, a final convolution layer (without batch norm and ReLU) is applied to obtain the final output feature with a shape of $1024\times1$, to make our \methodname compatible with both the point-based encoder and range map-based encoder simultaneously.

Note that both our point-based and range map-based encoders can process scene point clouds with a variable size of $K$. This makes our \methodname applicable to real-world scenarios such as autonomous driving datasets where the LiDAR sensor often collects scene point clouds with a different number of points $K$ at different frames, \emph{i.e.}, not a fixed $K$ across all frames.

\vspace{-0.2cm}
\subsection{Justification to Our Encoder Design}
\vspace{-0.1cm}

\textbf{Why not use more advanced encoders?} Although our \methodname has achieved promising performance, we are aware that the components of our \methodname such as the encoder might not be optimal as there could be many other choices to explore (\emph{e.g.}, replace the PointNet with PointNet++ \cite{Qi2017} or other recent point cloud processing techniques such as \cite{Wu2019, Lin2020, Wang20182}). However, we would like to emphasize that the main goal of this paper is not to exhaustively search for the best model (encoder, decoder, loss functions) for our \taskname task. Instead, our goal is, for the first time, to introduce this interesting new \taskname task and a new perception and forecasting pipeline \approachnameacronym which is built upon our \methodname and achieves competitive performance compared to the conventional pipeline. We hope that our \methodname only serves as a start, and future research in this direction can be encouraged and stronger neural network architectures to the proposed \taskname task can be proposed. 

\textbf{Why compress the feature in the encoder?} Since our proposed \taskname is a dense prediction task, \emph{i.e.}, predicting the future locations of $>100k$ points, it seems to be straightforward to use a fully convolutional architecture without feature compression (\emph{e.g.}, the pooling operator, a large stride in the convolution layer, or the graph pooling layer) in order to maintain fine-grained details in the feature space of the scene point cloud. This is also what we planned to implement at the beginning. However, we later realized that predicting a sequence of large-scale scene point clouds without feature compression can easily reach the GPU memory limits. Note that our sequences are very large (e.g., 30 frames) of large-scale point clouds with $>100k$ points. As a result, it is necessary to compress the intermediate feature to reduce the GPU memory requirement, \emph{e.g.}, in our two encoders, we compress the feature to a dimension of $1024$. We hope that future research in memory-efficient point cloud sequence processing techniques will be encouraged to further improve performance of the \taskname task without feature compression.

\vspace{-0.2cm}
\subsection{Justification to the Loss Function}
\vspace{-0.1cm}

As mentioned in Sec. 3.2 of the main paper, we use the Chamfer distance (CD) as our primary loss function for both the point-based and range map-based methods\footnote{Two additional losses are only applied for range map-based methods.}, which is also a standard practice in most prior works for object point cloud generation/completion \cite{Xia2019, Zou2020, Wang2020, Huang2020, Yuan2018, stypukowski2019}. Although a few works \cite{Yuan2018, Wu2020} also considered using the other well-known Earth mover distance (EMD) as the loss function during training, it is commonly accepted that computing the EMD is much more computationally expensive than computing CD, which is especially true for large-scale scene point clouds in our \taskname task. In fact, if we use the CD as our loss function, our \methodname for predicting future $10$ frames of scene point clouds will require about one day to train. However, if we use the EMD as our loss function, the training time for the same model increases to more than a week. Therefore, we only considered using the CD loss for training purpose due to its efficiency.

\vspace{-0.2cm}
\subsection{Other Implementation Details}
\vspace{-0.1cm}

We train the entire network \methodname in an end-to-end fashion using the Adam \cite{Kingma2015} optimizer for $30$ epochs with a learning rate of $1 \times 10^{-4}$, betas of $0.9$ and $0.999$, and batch size of $16$. For range map-based methods, we use $\lambda_1$ of $0.1$ and $\lambda_2$ of $0.1$ for the additional loss components in the Eq.~(1) of the main paper. For point-based methods, $\lambda_1$ and $\lambda_2$ are simply zeros. For temporal dynamics modeling in Sec. 3.2 of the main paper, we use a standard two-layer LSTM with the hidden feature dimension of $1024$.

\vspace{-0.3cm}
\section{End-to-End Perception and Trajectory Forecasting Evaluation}
\vspace{-0.2cm}
\subsection{Matching Step and Evaluation Metrics}
\vspace{-0.1cm}

As mentioned in Sec. 4 of the main paper, we match the predicted future trajectories and GT future trajectories using the Hungarian algorithm in order to establish the correspondences and compute the ADE/FDE metrics. Specifically, we use the pairwise ADE\footnote{Note that the pairwise ADE is different from the final ADE metric used in evaluation. The pairwise ADE is computed for one pair of GT and predicted trajectory by averaging the displacement error over all frames while the final ADE metric is computed by averaging the displacement error over all true positives and all frames.} as the metric for the matching. First, we compute the pairwise ADE between every possible pair of predicted and GT future trajectories. For example, if we have $U$ predicted and $V$ GT future trajectories, we will obtain a $U \times V$ matrix with each entry being the pairwise ADE for the pair of GT and predicted trajectories. Note that, for GT objects that do not have trajectories in all future $N$ frames, we mask out the missing frames and still use the available trajectories in valid frames to compute the pairwise ADE. 

Once we compute the pairwise ADE matrix with a shape of $U \times V$, we feed it to the Hungarian algorithm to obtain a one-to-one correspondence (\emph{i.e.}, a bijection) between the predicted and GT trajectories. As $U$ and $V$ may be not equal, there could be predicted trajectories or GT trajectories without having a correspondence. Also, for those predicted trajectories which have matched GT trajectories, we reject the matching if the corresponding pairwise ADE is higher than a threshold. In other words, for this predicted trajectory, the matched GT trajectory is the best option that the Hungarian algorithm can find and match it with, which however is still not good enough because the predicted trajectory and its matched GT trajectory have a large displacement error (a high pairwise ADE). As a result, we consider the predicted trajectories as true positives if and only if they have the corresponding matched GTs and also the corresponding pairwise ADEs are lower than the ADE threshold. The rest of unmatched (or matching rejected) predicted trajectories are considered as false positives and the unmatched (or matching rejected) GT trajectories are considered as false negatives. Then, based on the established correspondences for true positives, we can now compute the final ADE/FDE metrics as in the conventional modularized trajectory forecasting evaluation:
\vspace{-0.2cm}
\begin{equation}
    \text{ADE} = \frac{\sum_{i=1}^{|\mathcal{T}|} \text{ADE}_{\text{pairwise}}^i}{|\mathcal{T}|}, \ \mathrm{where}~ADE_{\text{pairwise}}^i = \frac{\sum_{j=1}^{N_{\text{valid}}^i} |\widetilde{\mathbf{x}}_j^i - \mathbf{x}_j^i|_2}{N_{\text{valid}}^i},
\end{equation}
\vspace{-0.2cm}
\begin{equation}
    \text{FDE} = \frac{\sum_{i=1}^{|\mathcal{T}|} \text{FDE}_{\text{pairwise}}^i}{|\mathcal{T}|}, \ \mathrm{where}~FDE_{\text{pairwise}}^i = |\widetilde{\mathbf{x}}_{N_{\text{valid}}^i}^i - \mathbf{x}_{N_{\text{valid}}^i}^i|_2,
\end{equation}
where $\mathcal{T}$ is the set containing all true positives and $|\mathcal{T}|$ denotes the number of true positives. Also, $N_{\text{valid}}^i$ denotes the valid number of frames for the pair of $i$th predicted and GT trajectory which can vary across true positives, $\mathbf{x}_j^i$ and $\widetilde{\mathbf{x}}_j^i$ denote the ground truth and predicted ground positions at the frame $j$ for the pair of the $i$th true positive.

Moreover, as we can compute the above ADE/FDE metrics at a given pairwise ADE threshold\footnote{A different threshold for the pairwise ADE corresponds to a specific recall value of the pipeline.}, we can actually compute the full ADE-over-recall and FDE-over-recall curves and ultimately compute the AADE/AFDE metrics through integration. To approximate the integration, we average the values of ADE/FDE at a maximum of $L$ recall values linearly distributed between $0\%$ and $100\%$ to compute the AADE/AFDE. We use $L$=$40$ in our experiments so that the interval of recall is $2.5\%$. Note that not ADE/FDE values at all $40$ recall values are used. In fact, we only integrate the ADE/FDE values up to the recall value that all methods can reach. This is because there is no ADE/FDE available for integration at a recall value that is beyond the maximum recall a method can reach. For example, in Fig. 3 (right) of the main paper, the recall that all three methods can reach is $0.425$, so we integrate the ADE/FDE to compute AADE/AFDE for all three methods at the recall values from $0.025$, $0.05$, $\cdots$, to $0.425$. Specifically, the AADE/AFDE metrics are computed as following:
\vspace{-0.2cm}
\begin{equation}
    \text{AADE} = \frac{1}{|\mathcal{R}|} \sum_{r \in \mathcal{R}} \text{ADE}_r,
\end{equation}
\begin{equation}
    \text{AFDE} = \frac{1}{|\mathcal{R}|} \sum_{r \in \mathcal{R}} \text{FDE}_r,
\end{equation}
where $\mathcal{R}$ is the set containing all the recall values at which the ADE/FDE are used for integration and $|\mathcal{R}|$ denotes the number of recall values. Again, in the example of Fig. 3 of the main paper, the recall set $\mathcal{R}=\{0.025$, $0.05$, $\cdots$, $0.425$\} and $|\mathcal{R}|=17$.

\vspace{-0.2cm}
\subsection{Alternatives of the Matching Step}
\vspace{-0.1cm}

In the above subsection, we have described the matching step used in our experiments in the main paper. Specifically, the matching is conducted between the predicted and GT trajectories in future $N$ frames based on the pairwise ADE values. Alternatively, there could be other variants of this matching step: (1) instead of performing the matching in future $N$ frames, we could match the trajectories from the tracking outputs and GT trajectories in past $M$ frames; (2) moreover, we could match the estimated trajectories (including both the tracking and forecasting outputs) and GT trajectories over the entire past and future $M+N$ frames; (3) instead of using the pairwise ADE for matching, we could compute the pairwise FDE matrix for matching. As a result, there are total $6$ different variants for matching, which we believe are all fair for comparison, although each variant might yield slightly different results. We will explore how different matching mechanism affects the performance in the future.

\vspace{-0.3cm}
\section{Experiments}
\vspace{-0.2cm}
\subsection{Datasets}
\vspace{-0.1cm}

We further clarify that our experiments (Table 1 and 3 of the main paper) on the KITTI and nuScenes datasets for all methods are conducted for the cars only, not including other objects so far, e.g., Cyclists, Pedestrians. As mentioned in the Sec. 5.1 of the main paper, we have conducted experiments on the nuScenes dataset only for the keyframes, despite the fact that there are many non-key frames. The reason that we only use the keyframes for training and evaluating our \methodname is that we want to reuse the trained \methodname in our \approachnameacronym pipeline. Since the nuScenes dataset only provides object labels for keyframes, our \approachnameacronym pipeline can only be evaluated on the keyframes. As a result, to reuse our \methodname in our \approachnameacronym pipeline, the easiest way is also to train \methodname for the keyframes, because otherwise the input frame rate might be different which will degrade performance of our \methodnamenosp.

\vspace{-0.2cm}
\subsection{Ablation Study}
\vspace{-0.1cm}

\begin{wraptable}{r}{0.38\linewidth}
\begin{center}
\vspace{-1.2cm}
\caption{Effect of the loss components.}
\vspace{-0.2cm}
\resizebox{0.38\textwidth}{!}{
\begin{tabular}{@{}cccrr@{}}
\toprule
$\mathcal{L}_{\text{cd}}$ & $\mathcal{L}_{1}$ & $\mathcal{L}_{\text{bce}}$ & CD$\downarrow$ & EMD$\downarrow$ \\ 
\midrule
\cmark &        & \cmark & 1.15 & 132.63 \\
\cmark & \cmark &        & 0.82 & 110.03 \\
\midrule
       & \cmark &        & 1.91 & 219.17 \\
       & \cmark & \cmark & 1.85 & 203.72 \\
\midrule
\cmark & \cmark & \cmark & 0.73 & 103.14 \\

\bottomrule
\end{tabular}}
\label{tab:loss}
\end{center}
\vspace{-0.6cm}
\end{wraptable}

To validate the design of our proposed \methodnamenosp, we conduct ablation study on the KITTI validation set. Also, since the baselines we have compared in Table 1 of the main paper may be weak, we believe that different variants of our \methodname can also serve as stronger baselines. Note that all experiments conducted in the ablation study are based on our range map-based \methodnamenosp.

\noindent\textbf{Effect of Loss Components.} To show that each component of our full object function in Eq. 1 of the main paper is effective, we turn on and off each loss component and check how the performance is affected. Note that the experiments are conducted in the 1.0s prediction setting. We summarize the results in Table \ref{tab:loss}. To start, the bottom row shows the performance of our full range map-based method on the KITTI val set. Then, to demonstrate that $\mathcal{L}_{\text{cd}}$ is useful, we turn it off and only use $\mathcal{L}_{1}$ and $\mathcal{L}_{\text{bce}}$ loss functions during training and the performance is shown in the second last row, where the CD and EMD metrics increase by a large margin. This demonstrates that using $\mathcal{L}_{\text{cd}}$ loss is crucial to the performance of our \methodname on the CD and EMD metrics. Moreover, we removed $\mathcal{L}_{\text{bce}}$ and saw that the CD and EMD metrics slightly increase in the third last row. We believe that this is because using the range mask can help remove some outliers in the generated scene point clouds. Additionally, to show that $\mathcal{L}_{1}$ loss is also necessary, we turn it off and the performance is shown in the top row. Compared to the last row, we can see that the CD and EMD metrics clearly increase (\emph{e.g.}, from $0.73$ to $1.15$ for CD), demonstrating that regularizing the intermediate range map output with $\mathcal{L}_{1}$ largely improves the quality of the generated scene point clouds. Similarly, we can turn $\mathcal{L}_{\text{bce}}$ off and the performance is shown in the second row. As a result, the performance for the CD and EMD metrics slightly decreases (\emph{e.g.}, from $0.73$ to $0.82$ for CD), although not by a large margin, showing that it is helpful to keep the $\mathcal{L}_{\text{bce}}$ in our \methodnamenosp.

\vspace{-0.3cm}
\bibliography{main}  